\documentclass[]{cit_lab_mfr}

\usepackage{hyperref}
\usepackage{cleveref}
\usepackage{verbatim}

\usepackage{wrapfig}
\usepackage{graphicx}
\usepackage{subcaption}
\usepackage{listings}
\usepackage{algorithm}
\usepackage{bm}
\usepackage{colortbl}
\usepackage{tocloft}

\usepackage{xcolor}
\usepackage[dvipsnames]{xcolor}
\usepackage[table]{xcolor}
\usepackage{multirow}
\usepackage{booktabs}
\usepackage{adjustbox}
\usepackage{tabularx}
\usepackage{siunitx}
\usepackage{enumitem}
\usepackage[table]{xcolor}
\usepackage{makecell}
\usepackage{pifont}
\usepackage{textgreek}
\usepackage{xspace}
\usepackage{tikz}
\usepackage{subcaption}
\usepackage{multicol}

\usepackage{tocloft}
\usepackage{amsmath}

\definecolor{cvprblue}{rgb}{0.21,0.49,0.74}
\definecolor{ForestGreen}{RGB}{34,139,34}
\definecolor{dino}{RGB}{249,231,227}
\definecolor{aliceblue}{rgb}{0.94, 0.97, 1.0}
\definecolor{MyGreen}{HTML}{92D050}
\definecolor{MyBlue}{HTML}{00B0F0}
\definecolor{MyOrange}{HTML}{FFC000}
\definecolor{MyRed}{HTML}{C00000}
\definecolor{myblue}{HTML}{2665B5}
\definecolor{myorange}{HTML}{E68727}
\definecolor{myred}{HTML}{D23D30}

\usepackage{subcaption} %
\usepackage[toc,page,header]{appendix}

\usepackage{minitoc}

\title{RePose: A Real-Time 3D Human Pose Estimation and Biomechanical Analysis Framework for Rehabilitation}

\author{%

\parbox{\textwidth}{\centering
Junxiao Xue$^{1}$$^{\dagger}$, Pavel Smirnov$^{1}$, Ziao Li$^{2}$, Yunyun Shi$^{3}$, Shi Chen$^{3}$, Xinyi Yin$^{4}$\\ 
Xiaohan Yue$^{5}$, Lei Wang$^{6}$, Yiduo Wang$^{4}$, Feng Lin$^{7,8}$, Yijia Chen$^{1}$, Xiao Ma$^{1}$ \\
Xiaoran Yan$^{1}$, Qing Zhang$^{1}$, Fengjian Xue$^{3}$, Xuecheng Wu$^{3,\S}$
}}

\affiliation{%
\parbox{\textwidth}{\centering\small
$^1$Zhejiang Lab, $^2$Northeastern University, $^3$Xi’an Jiaotong University \\[1mm]
$^4$Zhengzhou University, $^5$Dalian Minzu University, $^6$Nanjing University of Aeronautics and Astronautics \\[1mm]
$^7$Fuyao University of Science and Technology, $^8$Xianghu Lab
}}

\contribution[\S]{Corresponding author}
\contribution[\dagger]{Project Lead}

\abstract{
We propose a real-time 3D human pose estimation and motion analysis method termed RePose for rehabilitation training. It is capable of real-time monitoring and evaluation of patients'motion during rehabilitation, providing immediate feedback and guidance to assist patients in executing rehabilitation exercises correctly. Firstly, we introduce a unified pipeline for end-to-end real-time human pose estimation and motion analysis using RGB video input from multiple cameras which can be applied to the field of rehabilitation training. The pipeline can help to monitor and correct patients'actions, thus aiding them in regaining muscle strength and motor functions. Secondly, we propose a fast tracking method for medical rehabilitation scenarios with multiple-person interference, which requires less than 1ms for tracking for a single frame. Additionally, we modify SmoothNet for real-time posture estimation, effectively reducing pose estimation errors and restoring the patient's true motion state, making it visually smoother. Finally, we use Unity platform for real-time monitoring and evaluation of patients' motion during rehabilitation, and to display the muscle stress conditions to assist patients with their rehabilitation training.
}

\date{\today}
\checkdata[Corresponding]{\href{mailto:xuecwu@gmail.com}{xuecwu@gmail.com}}

\begin{document}
\maketitle

\addtocontents{toc}{\protect\setcounter{tocdepth}{-1}}

\section{Introduction}
\label{sec:intro}

Whether for post-operative recovery or treatment of various musculoskeletal disorders, participating in physical therapy and rehabilitation plans are usually crucial. However, making a clinician available for every rehabilitation session is an impractical task and also economically unreasonable \cite{1}. To address this issue, modern medical systems have begun adopting technologies and methods to provide remote monitoring and rehabilitation services. Remote monitoring, using sensors, mobile devices, and the internet, allows doctors to monitor patients' physiological indicators and rehabilitation progress in real-time, thus facilitating remote guidance and intervention. Such approaches can reduce the time patients stay in hospital and offer more convenient and cost-effective rehabilitation schemes.

Despite the development of various tools and devices to support physical rehabilitation, such as robot-assisted systems \cite{2}, virtual reality and gaming interfaces \cite{3}, and Kinect-based methods \cite{4}, there are still limitations in realizing a multi-functional and robust system for automatic monitoring and assessment of patient performance. Liao et al. \cite{5} introduced a deep learning framework for evaluating physical rehabilitation exercises, which assesses training through spatiotemporal modeling at multiple abstract levels. However, this method requires expensive optical motion capture systems for data acquisition. Swakshar et al. \cite{6} proposed a method based on Graph Convolution Networks to evaluate physical rehabilitation exercises, offering self-attention of body joints, but still reliant on RGB-D cameras for input data. For the assessment of Parkinson's disease movement severity, Lu et al. \cite{7} designed a posture-based evaluation system. Gu et al. \cite{8} introduced an interactive computer vision system for home physical therapy that can track human movement and provide analysis, but lacks an assessment of specific muscle exertion levels. Although progress has been made, challenges remain in developing multi-functional and robust automatic monitoring and evaluation systems. To aid patients in receiving guidance and feedback in home physical therapy, an ideal support system needs to overcome two difficult challenges: firstly, the system must be capable to accurately analyze patient movements during exercises in real-time, which means it needs high-precision pose estimation capabilities to accurately capture joint angles and positions. Secondly, the system also needs to provide timely and specific feedback and guidance. This includes real-time suggestions and adjustments based on the patient's performance, as well as showing the correct posture and movement demonstrations to the patient. This helps patients to correct incorrect postures and ways of moving, ensuring they receive effective training and rehabilitation effects at home.

To overcome the aforementioned challenges, this paper proposes a method for real-time posture recognition in rehabilitation exercises, providing real-time feedback and guidance. This method is based on RGB video inputs and accurately estimates the patient's three-dimensional body position, enabling motion analysis and evaluation during exercise~\cite{55}. This method can be widely applied in various environments, such as analyzing human behavior in public places through surveillance cameras, or assisting individual rehabilitation training through smartphone cameras. By utilizing computer vision and deep learning techniques, our method has the following advantages. Firstly, it can monitor and evaluate the patient's rehabilitation behavior in real-time, providing immediate feedback and guidance to help the patient perform the exercise correctly. This is crucial for the effectiveness of the rehabilitation process, as it can help the patient correct incorrect movements, adjust postures, and enhance the exercise efficiency. Secondly, in experiments, we found that the range of motion for patients in rehabilitation scenarios is relatively smaller compared to healthy individuals, resulting in noticeable jitters in the reconstructed body model. To address this issue, we utilize the SmoothNet~\cite{47} to reduce jitters. This filter can make the reconstructed body model smoother and more realistic, improving the accuracy and reliability of posture recognition. Furthermore, in rehabilitation scenarios with multiple participants (such as family members, doctors, etc.), the model only tracks and detects the behavior and movements of a specific individual rehabilitation patient. This helps focus on analyzing and evaluating the specific patient's rehabilitation progress, providing personalized rehabilitation plans and support. Finally, we utilize the Unity platform to visually demonstrate muscle and skeletal forces. By creating virtual environments and presenting them in three-dimensional forms, we can visualize the forces on muscles and skeletons during the rehabilitation process, enhancing the effectiveness and quality of the rehabilitation treatment experience.

To overcome the aforementioned challenges, this paper proposes a real-time posture recognition method for rehabilitation exercises that provides immediate feedback and guidance. Based on RGB images or video input, this method can accurately estimate the patient's three-dimensional body position, enabling the analysis and assessment of actions during the movement process. This method can be widely applied in various environments, such as analyzing human behavior in public places through surveillance cameras, or assisting personal rehabilitation training through smartphone cameras. By using computer vision and deep learning technologies~\cite{51, 54}, our method has the following advantages. Firstly, it can monitor and evaluate the patient's rehabilitation behavior in real-time, providing immediate feedback and guidance, helping patients perform movements correctly. This is vital for the effectiveness of the rehabilitation process, helping patients correct errors promptly, adjust posture, and enhance exercise effects. Secondly, 
we integrated SmoothNet into pipeline to reduce jitters. It can lower high-frequency components in the data, making the reconstructed human model more stable, thus enhancing the accuracy and reliability of posture recognition. Thirdly, in rehabilitation scenarios with multiple participants (such as family members, doctors, etc.), the model only tracks and detects the behavior and actions of a rehabilitation patient. This helps concentrate efforts on analyzing and evaluating the specific patient's rehabilitation progress, providing personalized rehabilitation plans and support~\cite{53}. Finally, we leverage the Unity platform to visually demonstrate the muscular and skeletal stresses. Through the creation of a virtual environment and its three-dimensional representation, we are able to visually display the forces exerted on the musculoskeletal system during rehabilitation, thus enhancing the effectiveness and quality of the therapeutic experience.

In conclusion, the main contributions of this paper are as follows:
\begin{itemize}
    \item We propose RePose, a real-time human pose estimation pipeline that maintains approximately 30 FPS on a Linux system in a multi-person medical rehabilitation scenario and is also capable of running on a Windows system.
    \item In the presence of multiple participants, our model exclusively tracks and detects the movements and actions of a specific individual undergoing rehabilitation. Additionally, a SmoothNet is utilized to reduce jitters caused by errors in pose estimation.
    \item By utilizing the Unity platform, we facilitate intuitive and real-time monitoring and assessment of rehabilitation behaviors and exhibit the muscular and skeletal stress conditions to assist patients in their rehabilitation training and therapy.
\end{itemize}

The organization of the paper is as follows: Section 2 provides an overview of related work. Section 3 introduces the proposed real-time rehabilitation exercise pose recognition framework and musculoskeletal stress analysis. Section 4 presents the validation of the proposed RePose framework on a public dataset. Section 5 proposes several promising directions for future research. Lastly, Section 6 concludes the paper.
\section{Related Work}
\label{sec:related_work}

\subsection{Human Pose Estimation}
Human Pose Estimation constitutes the process of locating human body parts and constructing human representations, such as the skeletal structure \cite{9}, from input data including images or videos. Human Pose Estimation (HPE) provides quantitative information regarding human motion, which can be utilized by medical professionals to diagnose complex conditions, create rehabilitative training programs, and guide physical therapy.

\textbf{Object Detection.} For a comprehensive image understanding, it is essential not just to classify different images but also to estimate the concepts and locations of objects contained within each image. General object detection aims to locate and classify objects present in any given image, primarily falling into two categories: one that follows the traditional object detection workflow, initially generating region proposals and then classifying each proposal into different object categories. The other views object detection as a regression or classification problem, adopting a unified framework to directly produce final results (both category and location) \cite{10}. Region proposal-based methods such as Fast R-CNN \cite{11}, Faster R-CNN \cite{12}, Feature Pyramid Networks (FPN) \cite{13}, and Mask R-CNN \cite{14} generate region proposals followed by object classification and localization based on these proposals. On the other hand, the regression/classification-based methods have adopted different object detection methodologies. They regard object detection as a regression or classification challenge, aiming to predict an object's category and location directly without depending on region proposals. The regression/classification-based methods include Deconvolutional Single Shot Detector (DSSD) \cite{15}, Deeply Supervised Object Detectors (DSOD) \cite{16}, YOLOv5 \cite{17}, and YOLOv8 \cite{18}. These methods typically extract features from input images using deep neural networks and employ various techniques (such as Anchor boxes) to predict the categories and bounding box coordinates of objects. While region proposal-based methods achieve high precision at a significant computational cost, regression/classification-based methods prioritize real-time processing at the expense of some accuracy.

\textbf{2D Human Pose Estimation.} The methodology of 2D Human Pose Estimation (2D HPE) involves estimating the two-dimensional spatial location of human body keypoints from images or video sequences. Depending on the number of subjects in the image, this can be categorized into two distinct approaches: single-person pose estimation and multi-person pose estimation~\cite{52}. Single-person pose estimation is a foundational research area within 2D HPE that does not require consideration of occlusions between targets. In general, single-person pose estimation can be approached via regression methods or heatmap-based methods. Regression methods employ an end-to-end framework to learn the mapping from input images to the positions of body joints or parameters of a human body model. By training a deep neural network, regression methods are capable of predicting the precise locations of body joints directly from images. For instance, the DeepPose method by Toshev et al. \cite{19} tackles the problem of body joint regression based on Deep Neural Networks (DNNs). Sun et al. \cite{20} introduced “Compositional Pose Regression,” a structure-aware regression method based on ResNet-50, utilizing a reparameterized pose representation that substitutes joints with bones. The advantage of these methods lies in their ability to estimate the positions of multiple joints simultaneously and to handle variations in human body shapes and postures. Another prevalent approach is the heatmap-based method. This approach aims to predict the approximate locations of body parts and joints, supervised by their representations on heatmaps \cite{21}. Specifically, heatmap-based HPE methods estimate 2D heatmaps instead of directly estimating the two-dimensional coordinates of body joints by placing 2D Gaussian kernels at each joint's location. Heatmaps are two-dimensional matrices of the same size as the image, where each element indicates the confidence level for the corresponding body part or joint. By training a deep neural network, heatmap-based methods are able to generate heatmaps from the input image and extract positional information of keypoints from them. For example, Newell et al. \cite{22} proposed the ``Stacked Hourglass'' encoder-decoder network that performs iterative bottom-up and top-down processing with intermediate supervision. Sun et al. \cite{23} introduced a High-Resolution Net (HRNet) that acquires more accurate keypoint heatmap predictions by learning reliable high-resolution representations through the parallel connection of multi-resolution subnetworks and repeated multi-scale fusions.

\textbf{3D Human Pose Estimation.} The objective of 3D Human Pose Estimation (3D HPE) is to predict the position of human body joints within three-dimensional space. The aim of 3D HPE is to accurately estimate the three-dimensional coordinates of the human body's joints from a given input, such as an image or video sequence. This process necessitates an understanding of the spatial relationships between different body parts and inferring their positions within the three-dimensional world. Precise estimation of 3D human pose can provide valuable information about body mechanics, which can be utilized in various applications including motion analysis, rehabilitation, and animation. Approaches to 3D HPE can be generally categorized into two types: model-based and model-free. Model-based methods utilize prior knowledge about human anatomy and kinematics to reconstruct a three-dimensional posture, typically by fitting a parametric human body model to the input data and optimizing its parameters to align with observed joint positions. On the other hand, model-free methods employ deep learning techniques to directly regress the three-dimensional joint positions from input data, without explicitly relying on a predefined human body model. Based on the number of views and the number of individuals in the scene, 3D HPE methods are subdivided into three categories: single-view single-person 3D HPE, single-view multi-person 3D HPE, and multi-view 3D HPE. In this paper we primarily investigate multi-view and model-free methods.  Iskakov et al. \cite{42} proposed a multi-view 3D human pose estimation scheme based on a learnable triangulation method. Zhang et al. \cite{29} introduced AdaFuse, which aggregates 2D keypoint heatmaps from multi-view images into a three-dimensional structural model, based on all camera parameters after calibration. Tu et al. \cite{30} presented VoxelPose, which aggregates features from each camera view in a 3D voxel space. They designed a cuboid proposal network and a pose regression network to localize all individuals and estimate their 3D poses, respectively. Zhang et al. \cite{31} proposed the Multi-view Pose transformer (MvP), a model that regresses 3D poses directly from multi-view images without reliance on any intermediary tasks. To reduce the computational cost of VoxelPose, Ye et al. put forward Faster VoxelPose \cite{32}, which reprojects the feature volume onto three two-dimensional coordinate planes and estimates the X, Y, and Z coordinates separately, thereby increasing the speed of VoxelPose by a factor of ten.

\subsection{Musculoskeletal Movement foress}
Human movement is driven and supported by the musculoskeletal system, which includes bones, skeletal muscles, joints, and soft connective tissues, together forming a complex multi-body system \cite{33}. To address the motion issues within musculoskeletal models, both forward and inverse solutions are typically employed. Forward kinematics approaches take electromyography (EMG) measurements \cite{34,35} or neural commands obtained from controllers \cite{36,37} as model inputs to predict the motion state of the human body. This method can be used to estimate the intensity and timing of muscle activities, thereby inferring the body's posture and movement. However, forward kinematics approaches are relatively computationally inefficient and challenging to apply to complex motion studies such as three-dimensional gait analysis. In contrast, inverse kinematics approaches use measured joint kinematics and external forces as model inputs to estimate joint torques and muscle forces~\cite{38,39}. This method infers muscle activation patterns and force outputs by back-calculating muscle torques and joint forces, and is significantly valuable in studying human motion control and rehabilitation therapy.

\subsection{Rehabilitation Systems}
Various home-use devices for rehabilitation have been developed, including robots, wearable devices, and Kinect-based rehabilitation applications. Robotic rehabilitation~\cite{48} devices can assist patients in their rehabilitation exercises through the robot's strength and precision. These devices often feature multiple joints and sensors to offer personalized rehabilitation therapy. For instance, lower-limb rehabilitation robots like Lokomat, Hocoma, and others, utilize principles of extracorporeal therapy by supporting the patient's lower limb movements to facilitate the rehabilitation of lower limb dysfunctions. Robotic rehabilitation devices provide highly accurate motion control, which can precisely monitor and correct patients' actions, thereby aiding in the recovery of muscle strength and motor function. However, these robots~\cite{49, 50} are costly, bulky, and only offer specific types of rehabilitation training. Wearable devices are directly worn on the body to monitor and assist the patient's movements. These devices are compact and portable, allowing patients to wear them at any time and place for rehabilitation exercises. Nonetheless, wearable devices usually only monitor and support specific parts or types of movement. For whole-body movements or rehabilitation involving multiple joints, additional equipment or methods may be required. Kinect-based rehabilitation applications utilize the Microsoft Kinect sensor for rehabilitation training. Kinect-based applications do not require direct patient contact with the device and can support various types of rehabilitation exercises by using the camera to identify and track the patient's movements. However, due to the limitations of camera technology, they may not provide highly accurate motion tracking and control, and the purchase of Kinect-specific hardware is necessary.
\section{Method}

\begin{figure}
    \centering
    \includegraphics[width=0.8\linewidth]{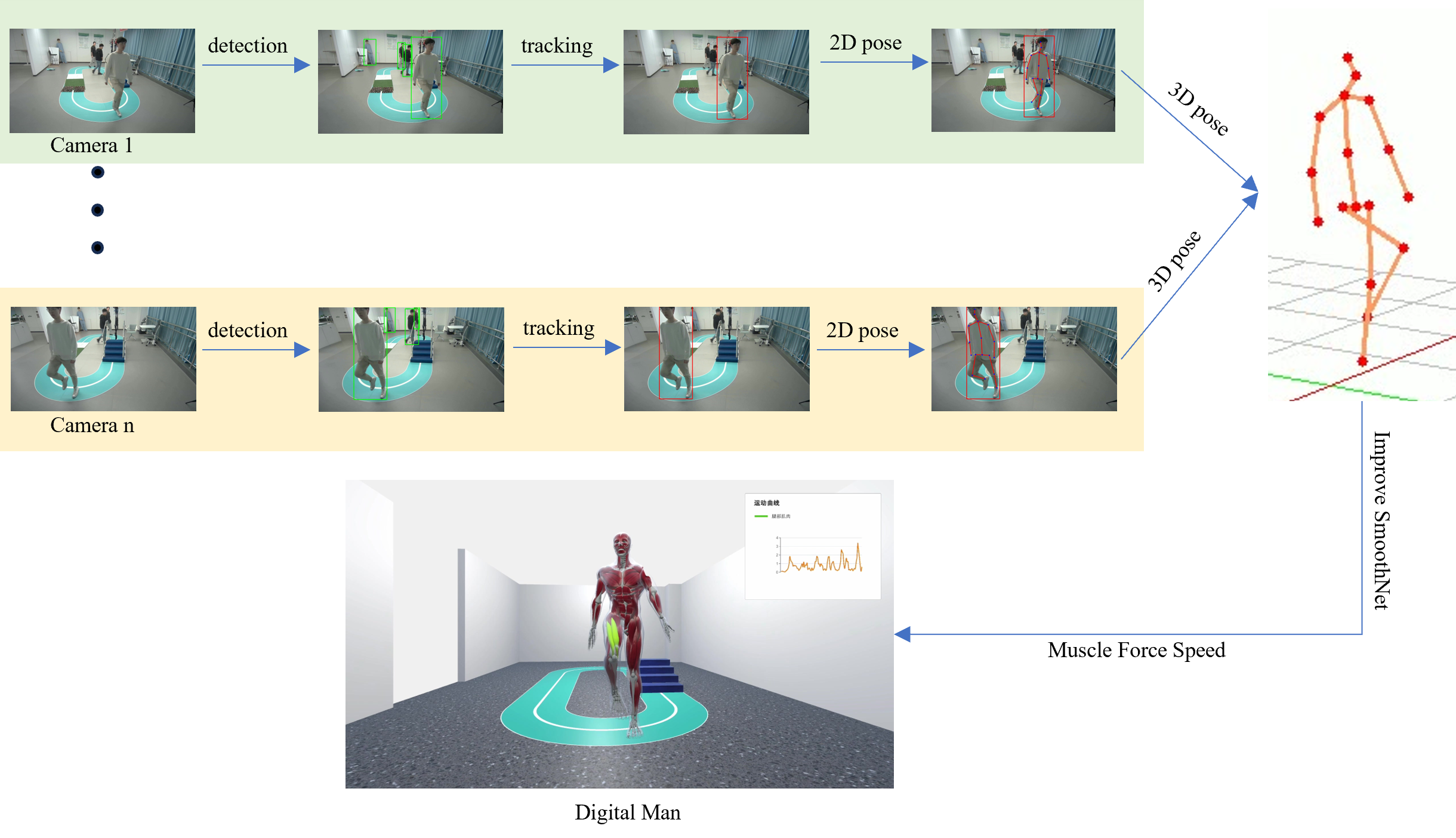}
    \caption{Overview of introduced RePose multi-camera real-time digital human pipeline. The red box represents the tracking box. Each camera generates a 2D pose and synthesize a 3D pose. And then generates to digital human.}
    \label{Figure 1}
\end{figure}

\textbf{Solution Framework.} Our approach reconstructs a 3D virtual digital human body model from multiple video streams. Prior to 3D HPE, it is necessary to calibrate the intrinsic and extrinsic parameters of the cameras used to acquire the mapping relationship between the 2D image coordinate system and the 3D spatial coordinate system. This calibration allows to convert detected human 2D coordinates into 3D coordinates. Subsequently, multi-camera system capture multiple synchronous video streams, and the following processing steps are applied to each frame of each video stream. The YOLO-v5~\cite{17} object detection model is employed to estimate the position and output bounding boxes of people in video. After that, tracking algorithm is applied to track the person whose pose should be estimated. For the multi-view 3D pose estimation, we follow the approach proposed in \cite{46}. We crop an image of the person of interest using tracked bounding box and estimate the positions of 17 human body joints for each frame. Based on the 2D coordinates of the 17 detected joints in multiple image frames at the same timestamp, the positions of the corresponding joints in 3D space are estimated using the learnable linear triangulation method. Next, we use SmoothNet to filter the 3d joint positions and feed them into the digital human part for musculoskeletal calculations. The process flow is illustrated in Figure \ref{Figure 1}.

\subsection{Human Pose Estimation Model}
\vspace{0.3pt}\noindent\textbf{Object Detection.}\hspace{1ex}
In the object detection phase, OpenCV library is utilized to read video streams from cameras, and pretrained YOLOv5 model to detect people. Briefly the process can be described as follows. Firstly, the backbone network extracts features from the input image; then, the feature fusion module outputs feature maps of different scales; finally, the prediction module outputs regression parameters for target boundary boxes and target categories based on the feature maps. Identified human figures are extracted by setting non-maximum suppression (NMS).

\vspace{0.3pt}\noindent\textbf{Object Tracking.}\hspace{1ex}
In actual rehabilitation scenarios, it is common to have situations where either an individual or multiple people are present simultaneously. Here, an individual refers to the patient whose pose should be estimated and multiple people can include the patient, family members, and medical personnel. In such scenarios, it is necessary to track the patient in order to better monitor and assist the rehabilitation process. When there is only a single person present, that is, n=1, real-time individual tracking is conducted directly. However, if multiple human figures are detected, resulting in n bounding boxes $Bboxs = [B_0,B_1...B_n]$, that is, n$>$1, it becomes necessary to employ a tracking algorithm to focus specifically on the rehabilitation control and assessment of the patient.

\begin{figure}
    \centering
    \includegraphics[width=0.4\linewidth]{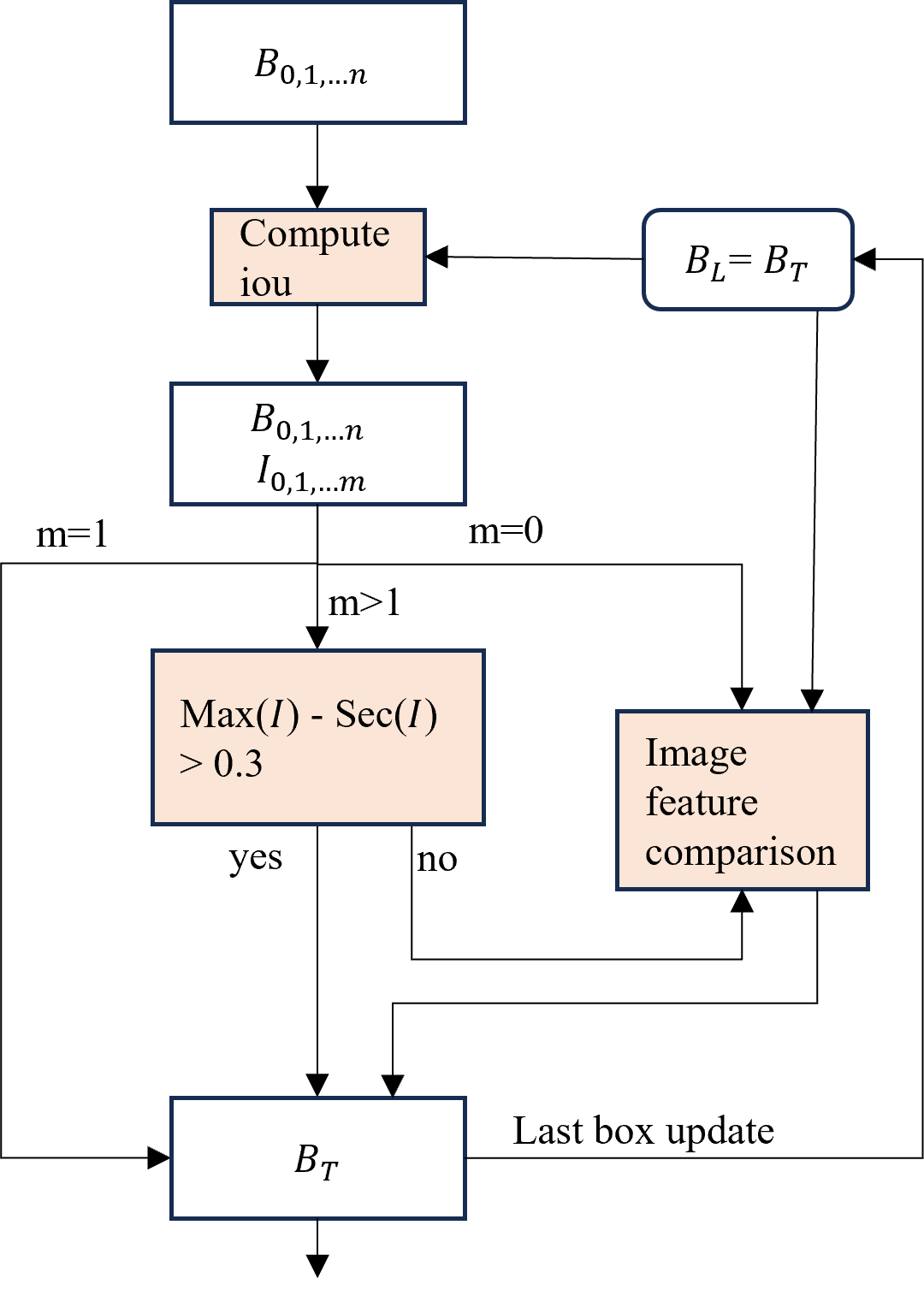}
    \caption{Flowchart of the proposed real-time single-person tracking algorithm with hierarchical decision logic designed for complex multi-person rehabilitation scenarios.}
    \label{Figure 2}
    \vspace{-0.5em}
\end{figure}

As shown in Figure \ref{Figure 2} above, we employ the similarity calculation based on the Intersection over Union (IoU) to measure the degree of overlap between two bounding boxes. The larger the overlapping area, the larger the value of IoU. Specifically, first we save the target box $B_L$ of the previous frame as the tracking object, and then calculate its IoU with all the boxes in the current frame Bboxs. If the IoU is greater than a certain threshold (default value is 0.2), the $B_i$ is considered a candidate box, and the number of candidate boxes is m (m$<=$n). If m=1, the current candidate box is considered to be the best tracking target $B_T$. If m$>$1, we sort the candidate boxes and calculate the difference between Bboxs with highest values of IoU and if the difference is greater than or equal to 0.3, the Bbox with the largest value of IoU is considered to be the best tracking target $B_T$. 
Otherwise (or for m=0), indicating that the tracking box $B_T$ cannot be accurately located solely based on IoU, some other image feature extraction methods are used to estimate similarity of detected Bboxs for the current frame and tracked Bbox from the previous frame. Specifically, the color features of the candidate n Bboxes are calculated, and the same method is applied to the $B_L$. Then, feature similarity between each candidate $Bboxs$ and the $B_L$ is estimated one by one and the bounding box with the highest value of feature similarity is chosen as the tracking object $B_T$.

Although other existing solutions for tracking can provide higher accuracy, they would require more computational resources. However in the actual medical rehabilitation scenarios, there is generally no overlap between the patient and the surrounding people, such as, m is generally equal to 1. For such case our tracking algorithm does not require feature calculations, it only relies on IOUs comparison, which results in high processing speed and high accuracy. As shown in Table \ref{tab1}, the average inference time of our tracking algorithm takes less than 1 millisecond.

\vspace{0.3pt}\noindent\textbf{2D Pose Estimation.}\hspace{1ex}
Our pose estimation algorithm follows the approach proposed in \cite{46}. Initially, we use the bounding box Bbox to crop the image of the patient $I_c$. Subsequently, 2D pose estimation backbone ResNet152 takes this image as an input to estimate joint heatmaps $H_{c,j}=h_\theta (I_c )_j$, where $H_{c,j}$ represents the heatmap for joint j in image $I_c$,  $h_\theta$ denotes a 1×1 convolution kernel, $j \in 1,….,J$, and $J$ is the number of joints. To estimate the 2D positions based on the joint heatmaps, we first compute the softmax along the spatial domain of the heatmap as follows:

\begin{equation}
H_{c,j}' = \frac{\exp(\alpha H_{c,j})}{\sum_{r_x=1}^W \sum_{r_y=1}^H \exp(\alpha H_{c,j}(r))}
\end{equation}

Here, $H_{c,j}'$ indicates the heatmap after applying the softmax operation, W and H respectively denote the width and height of the heatmap, $H_{c,j}=h_\theta (I_c)_j$ refers to the heatmap value at coordinates r, and $\alpha$ is a hyperparameter that adjusts the heatmap. The central position for each joint in 2D is calculated using the soft-argmax operation on the heatmap $H_{c,j}'$:

\begin{equation}
x_{c,j} = \sum_{r_x=1}^W \sum_{r_y=1}^H r\cdot (H_{c,j}' (r)),
\end{equation}
where $x_{c,j}\in R^{J×2}$ denotes the two-dimensional coordinates corresponding to the joint points in the heatmap $H_{c,j}' (r)$, thereby obtaining the two-dimensional coordinates of all joint points in each view.

\vspace{0.3pt}\noindent\textbf{3D pose estimation.}\hspace{1ex}
After obtaining the two-dimensional joint positions $x_{c,j}$ in the image, we infer the three-dimensional positions of the joints $y_j$ using linear algebra triangulation. To facilitate computation, the problem of computing three-dimensional coordinates is reduced to solving an overdetermined system of equations on the homogeneous three-dimensional coordinate vector:
\begin{equation}
(w_j \circ A_j ) y_j=0.
\end{equation}
Here, $y_j$ represents the sought three-dimensional coordinate vector, $A_j\in R^{2C×4}$ is a matrix composed of the projection matrix $P_C$ and the two-dimensional joint coordinates $x_{c,j}$. $w_j$ is estimated by the convolutional network $q_\phi$, with inputs from the output of the ResNet152 network, where $\circ$ denotes the Hadamard product. Finally, the three-dimensional positions of the joints $y_j$ are solved using Singular Value Decomposition (SVD).

\vspace{0.3pt}\noindent\textbf{Modify SmoothNet.}\hspace{1ex}
To attain smoother 3D pose estimations and achieve higher frames per second (FPS), we have refined the SmoothNet~\cite{47} algorithm for real-time 3D pose estimation. The original SmoothNet algorithm is only applicable to video data, requiring temporal information not only from current and previous frames but future frames as well. However for real-time processing only historical data is available. Furthermore, our goal was also to enhance the visual coherence of outputs through interpolation of estimated poses between frames, thereby increasing the FPS.
\begin{figure}
    \centering
    \includegraphics[width=0.5\linewidth]{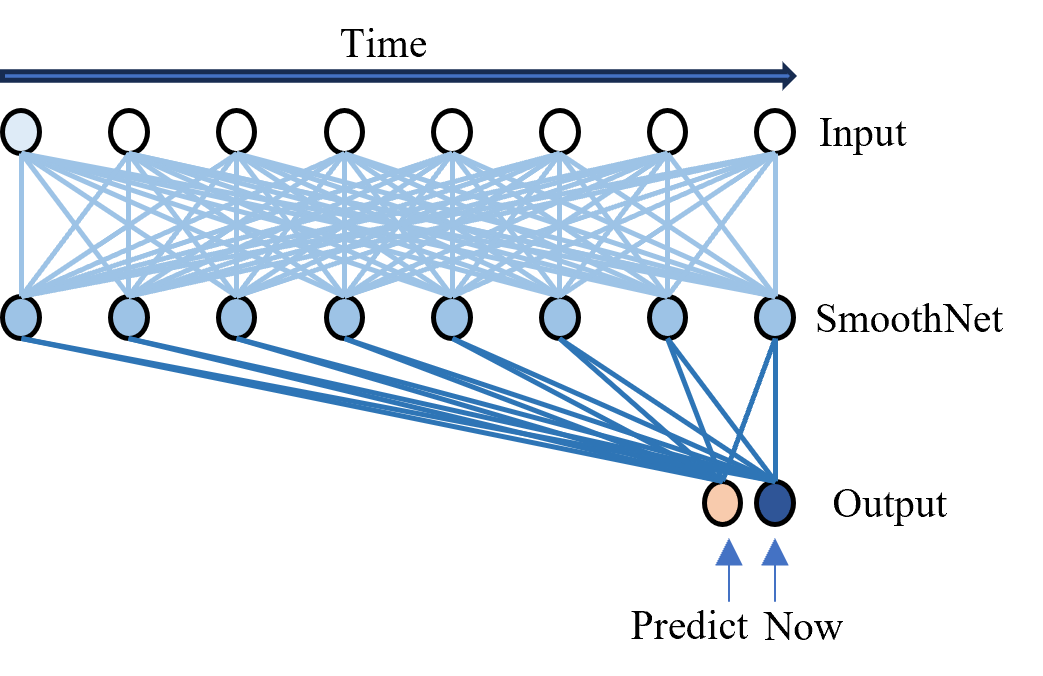}
    \caption{Architecture of the modified SmoothNet for real-time temporal refinement and motion prediction.}
    \label{Figure 3}
\end{figure}
The model architecture follows the original SmoothNet, with modifications to the input and output data as well as the training loss. Figure \ref{Figure 3} illustrates the schematic of the Modify SmoothNet algorithm. In the Modify model, we input the 3D poses of the current and preceding eight frames, enabling the model to output poses for two frames, the current frame and the predicted intermediate frame. This design allows us to double the FPS for smoother visualization.

Moreover, beyond using position loss, we have incorporated velocity loss to ensure a smoother output of poses. We use velocity loss since we output only two joint points, i.e., the current frame and the predicted frame. The training loss (Loss) function is computed as:

\begin{align}
L_{total} &= \alpha L_{pos}+(1-\alpha) L_{acc}, \\
L_{pos}   &= \frac{1}{T \times C} \sum_{t=0}^T \sum_{i=0}^C |G_{t,i} - Y_{t,i}|, \\
L_{vel}   &= \frac{1}{T \times C} \sum_{t=0}^T \sum_{i=0}^C |G_{t,i} - G_{t-1,i}||Y_{t,i} - Y_{t-1,i}|.
\end{align}

Here, $\alpha$ denotes the proportionality coefficient between the two losses, T and C represent the number of output time frames and output joints, respectively, $G_{t,i}$ and $Y_{t,i}$ represent the ground truth joint positions and the estimated joint positions, while $|G_{t,i} - G_{t-1,i}|$ and $|Y_{t,i} - Y_{t-1,i}|$ signify the velocity of joint i at time frame t.

\subsection{Calculate muscle forces}
\label{sec:training}

\begin{figure}
    \centering
    \includegraphics[width=0.4\linewidth]{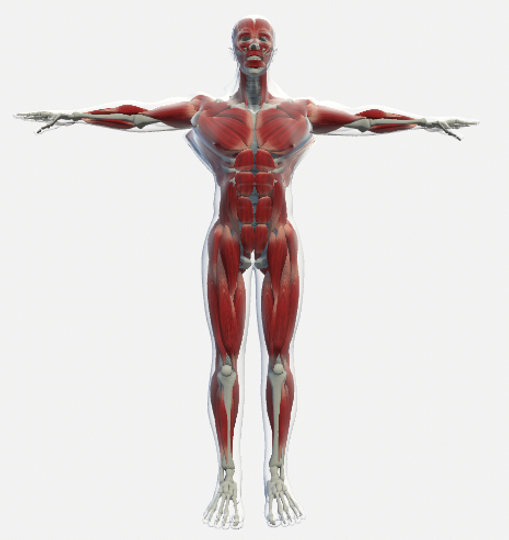}
    \caption{The musculoskeletal digital human model used for muscle stress quantification and visualization.}
    \label{Figure 4}
\end{figure}

\vspace{0.3pt}\noindent\textbf{Calculation of muscle stress quantification.}\hspace{1ex}
The world coordinates derived from the above models are based on a right-handed coordinate system, whereas Unity commonly employs a left-handed coordinate system. Consequently, we first transform the three-dimensional coordinate system using the Rodrigues' rotation formula to align the origin and axes with those of Unity's coordinate system. Subsequently, we utilize the FABRIK (Forward And Backward Reaching Inverse Kinematics) algorithm~\cite{40} to achieve inverse kinematics. Given the received coordinate data as input for target points, we drive the motion of a digital human model through the FABRIK algorithm, which calculates the positions and poses of the model’s joints based on the target point location and constraint conditions. We employ an original digital human model obtained from the Unity library, as depicted in Figure \ref{Figure 4}. By adjusting the digital human model's pose frame-by-frame, we ensure it aligns with the target points.

The FABRIK algorithm is based on the iterative solving approach, which approximates the target point through multiple iterations. The input consists of the target position t and the distances between each joint position
\begin{equation}
 d_i=|p_{i+1}-p_i |,
\end{equation}
where $p_i$ represents the joint position coordinates, i=1,…,n-1. The final joint positions $p_i$ are computed by iterating the algorithm.

Assuming $p_1$ is the root joint, the distance between the root joint and the target position is given by
\begin{equation}
dist=|p_1-t|.
\end{equation}

\begin{figure}
    \centering
    \includegraphics[width=0.8\linewidth]{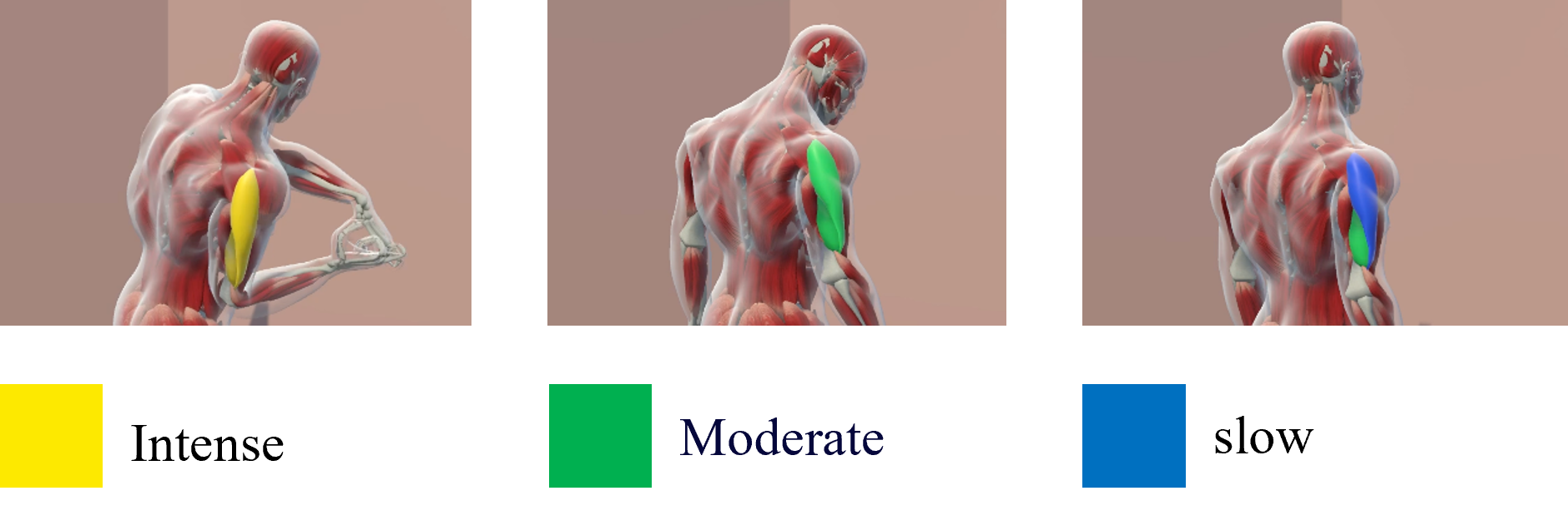}
    \caption{Color-coded heatmap representation for muscle stress visualization during rehabilitation exercises.}
    \label{Figure 5}
\end{figure}

If $dist>d_1+d_2+\dots +d_{n-1}$, then the target position is unreachable. In this case, the distance between the target position and the joint positions $r_i$ is determined using the following formula:

\begin{align}
r_i &= |t-p_i |, \\
\lambda_i &= \frac{d_i}{r_i}.
\end{align}

The new joint position $P_{i+1}$ is determined as:
\begin{equation}
p_{i+1}=(1-λ_i ) p_i+λ_{it}.
\end{equation}

If $dist \le d_1+d_2+\cdots+d_{n-1}$, then the target position is reachable. Assuming the initial position of the root joint $p_1$ is b, we first check if the distance between the end effector $p_n$ and the target t is greater than a tolerance value tol
\begin{equation}
{dif}_A=|p_n-t|.
\end{equation}

If ${dif}_A>tol$, we perform forward reaching by setting the end effector $p_n$ to the target position t, i.e., $p_n=t$. The distance between the target and joint positions $r_i$ is computed as:

\begin{align}
r_i &= |p_{i+1}-p_i |, \\
\lambda_i &= \frac{d_i}{r_i}.
\end{align}

The new joint position $p_i$ is determined as:
\begin{equation}
p_i=(1-λ_i ) p_{i+1}+λ_i p_i.
\end{equation}

Next, we perform backward reaching by setting the root joint $p_1$ to the initial position $p_1=b$. The distance between the new joint positions $r_i$ is computed as:

\begin{align}
r_i &= |p_{i+1}-p_i |, \\
\lambda_i &= \frac{d_i}{r_i}.
\end{align}

The final joint position is determined as:
\begin{equation}
p_{i+1}=(1-λ_i ) p_i+λ_i p_{i+1}.
\end{equation}

Throughout the algorithm's execution, FABRIK initiates by setting the initial joint positions of the digital human model. Subsequently, it iteratively adjusts the positions of the joints to bring the extremities of the digital human model, such as the hands or feet, as close as possible to the target point. Concurrently, the FABRIK algorithm adheres to the joint constraints, ensuring the digital human model's posture remains within physiological limits. This method allows for precise control over the digital human model, enabling it to move in response to the target data received.

\vspace{0.3pt}\noindent\textbf{Muscle Force Speed.}\hspace{1ex}
The calculation of muscle force typically involves the measurement of changes in velocity. When muscles are subjected to external forces or their own contraction, they generate movement, which, in turn, causes changes in velocity at related skeletal joints. By measuring the changes in velocity, we can infer the state of muscle force. Initially, it is essential to determine the correlation between muscle mesh skinning and skeletal joints. For instance, based on anatomical data, the triceps brachii are located on the posterior side of the arm, acting on the shoulder joint and forearm, and antagonizing the biceps brachii. Utilizing Unity software, one can identify the skeletal joints associated with the triceps brachii and use the velocity of these joints to deduce the muscle's force generation. Subsequently, the motion velocity of the skeletal joints is computed. At every frame update, the three-dimensional vector of the current frame is subtracted from that of the previous frame, and then divide it by the frame time difference (0.02f), resulting in a three-dimensional vector representing the velocity of the skeletal node, defined as:

\begin{equation}
V=\frac{(x_n,y_n,z_n ) - (x_{n-1},y_{n-1},z_{n-1} )}{0.02f}.
\end{equation}

Using coroutines allows for the continuous calculation of the joint's velocity vector. Since coroutines operate in the background, they can perform continuous calculations and updates without blocking the main thread during frame-by-frame execution. Finally, the muscle's color is adjusted in real-time based on changes in the skeletal motion velocity. The absolute value of the velocity vector is obtained and weighted on the x, y, and z axes to estimate the muscle's force intensity. This intensity is classified into three levels: intense, moderate, and slow, with thresholds defined as ``speed''. An intensity level is considered intense if ``speed'' is equal to or greater than 0.2, moderate if ``speed'' is less than 0.2 but greater than 0.08, and slow if ``speed'' is equal to or less than 0.08. These intensity levels are visually represented on the corresponding muscle mesh skinning using yellow, green, and blue colors, respectively.

\section{Experiment}
\subsection{Datasets and Evaluation Metric}

We adopt the most commonly used evaluation protocols. Protocol 1 is the Mean Per Joint Position Error (MPJPE) which measures the mean Euclidean distance between the ground truth and estimated joints in millimeters. The MPJPE formula is defined as:
\begin{equation}
E_{\text{MPJPE}}(f, S) = \frac{1}{N_S} \sum_{i=1}^{N_S} \left\| P_{f,S}^{(f)}(i) - P_{\text{gt},S}^{(f)}(i) \right\|_2,
\end{equation}
where f denotes a frame and S denotes the corresponding skeleton. $P_{f,S}^{(f)}(i)$  is the estimated position of joint i and $P_{gt,S}^{(f)} (i)$ is the corresponding ground truth position. All joints are considered, $N_S=17$. The MPJPE are averaged over all frames. Protocol 2 is the MPJPE after aligning the predicted 3D pose with the ground truth using translation, rotation, and scale (P-MPJPE).

{\renewcommand{\arraystretch}{1.2}
\setlength{\tabcolsep}{5pt}
\begin{table}[t]
\centering
\small
\caption{Real-time pipeline inference speed. ``Prep in detection'' refers to the image transformation prior to target detection, ``prep in 3d pose'' pertains to image cropping preceding the 3D segment, and ``W/O or W in Fps'' signifies whether fps is doubled through SmoothNet.}
\label{tab1}
\begin{tabular}{c|l|cc|c|ccc|c|cc}
\toprule
\multirow{2}{*}{\textbf{Num cam}} & \multirow{2}{*}{\textbf{Video}} & \multicolumn{2}{c|}{\textbf{Detection}} & \multirow{2}{*}{\textbf{Track}} & \multicolumn{3}{c|}{\textbf{3D Pose}} & \multirow{2}{*}{\makecell{\textbf{Total}\\\textbf{(ms)}}} & \multicolumn{2}{c}{\textbf{FPS}} \\
\cmidrule(lr){3-4} \cmidrule(lr){6-8} \cmidrule(lr){10-11}
& & Prep & YOLO & & Prep & 2D & 3D & & W/O SN & W SN \\
\midrule
4 & Offline & 9.1 & 2.1 & 2.5 & 9.3 & 6.0 & 1.7 & 30.7 & 33 & 66 \\ 
2 & Offline & 4.4 & 1.3 & 1.9 & 4.8 & 2.9 & 1.4 & 16.7 & 60 & 120 \\ 
4 & Online  & 25.0 & 2.7 & 2.1 & 13.2 & 5.7 & 4.9 & 53.6 & 19 & 38 \\
2 & Online  & 22.5 & 2.0 & 1.4 & 7.0 & 3.8 & 4.0 & 40.7 & 25 & 50 \\
\bottomrule
\end{tabular}
\end{table}}

To validate the accuracy and real-time performance of our modeling framework, we conduct validation tests at the Human3.6 Dataset \cite{41}. The Human3.6 Dataset have 4 high-resolution progressive scan cameras to acquire video data at 50 Hz. The dataset consists of 11 different human movement categories covering a wide range of movements in daily life, such as walking, running, lifting weights, etc. Each movement category has several different instances of human motion captured from different angles and views. Following previous works, we train a single model on five subjects (S1, S5, S6, S7, S8) and test it on two subjects (S9 and S11).

\subsection{Implementation Details}
Our pipeline is implemented on PyTorch. During training, for the 2D pose estimation network, we adhere to the setup as prescribed in \cite{46}. For YOLO-v5m, we employ the pre-trained data provided by the official source. For SmoothNet, we follow the training configuration detailed in \cite{47}, utilizing the human3.6 dataset for training purposes. Regarding model deployment, pipeline inference is conducted on a Linux server, while visualization is carried out on a Windows server utilizing the Unity platform. Notably, for the YOLO and 2D model components, we convert them into 16-bit floating-point type to diminish model weight, and employ TensorRT for model acceleration.

{\renewcommand{\arraystretch}{1.1}
\setlength{\tabcolsep}{3pt}
\begin{table*}[t]
\centering
\small
\caption{Quantitative comparison with the state-of-the-art methods on Human3.6M under Protocol 1. * means the ground truth bounding boxes provided by the dataset, $\dagger$ means bounding boxes with our proposed detection and tracking algorithm.}
\label{tab2}
\begin{tabular}{c|cccc|>{\columncolor{aliceblue}}c}
\toprule
\textbf{Methods} & \textbf{VideoPose3D}*~\cite{42} & \textbf{Anatomy3D}*~\cite{43} & \textbf{MHFormer}*~\cite{44} & \textbf{PoseFormerV2}*~\cite{45} & \textbf{Pipeline4}$\dagger$ (Ours) \\
\midrule
Dir.     & 45.16 & 41.49 & 39.17 & 41.26 & 25.61 \\ 
Dis.     & 46.67 & 43.80 & 43.06 & 45.45 & 33.44 \\ 
Eat      & 43.31 & 39.77 & 40.07 & 41.47 & 35.51 \\ 
Gre.     & 45.61 & 43.12 & 40.94 & 44.04 & 31.47 \\ 
Phon.    & 48.10 & 46.18 & 44.95 & 46.67 & 63.73 \\ 
Pho.     & 55.13 & 52.54 & 51.16 & 53.75 & 28.62 \\ 
Pose     & 44.62 & 42.16 & 40.63 & 42.63 & 25.44 \\ 
Pur.     & 44.28 & 41.85 & 41.29 & 42.64 & 29.68 \\ 
Sit      & 57.30 & 54.15 & 53.56 & 55.19 & 137.46 \\ 
SitD.    & 65.79 & 60.70 & 60.37 & 64.62 & 111.77 \\ 
Sm.      & 47.09 & 45.49 & 43.71 & 45.71 & 46.28 \\ 
Wait     & 43.99 & 41.56 & 42.12 & 42.89 & 26.70 \\ 
WD.      & 48.96 & 46.04 & 43.84 & 45.76 & 30.67 \\ 
Walk     & 32.80 & 31.41 & 29.76 & 32.34 & 27.72 \\ 
WT.      & 33.89 & 32.45 & 30.59 & 32.89 & 26.24 \\ 
\midrule
\textbf{Avg.} & 46.85 & 44.18 & 42.95 & 45.15 & 45.36 \\ 
\bottomrule
\end{tabular}
\end{table*}}

{\renewcommand{\arraystretch}{1.1}
\setlength{\tabcolsep}{6pt}
\begin{table*}[t]
\centering
\small
\caption{Quantitative comparison with the state-of-the-art methods on Human3.6M under Protocol 2.}
\label{tab3}
\begin{tabular}{c|cccc|>{\columncolor{aliceblue}}c}
\toprule
\textbf{Methods} & \textbf{VideoPose3D}*~\cite{42} & \textbf{Anatomy3D}*~\cite{43} & \textbf{MHFormer}*~\cite{44} & \textbf{PoseFormerV2}*~\cite{45} & \textbf{Pipeline}$\dagger$ \textbf{(Ours)} \\
\midrule
Dir.     & 34.12 & 33.01 & 31.36 & 32.34 & 23.38 \\ 
Dis.     & 36.12 & 35.29 & 34.91 & 35.87 & 31.27 \\ 
Eat      & 34.43 & 32.63 & 32.81 & 33.78 & 37.66 \\ 
Gre.     & 37.22 & 35.37 & 33.85 & 35.81 & 23.93 \\ 
Phon.    & 36.39 & 35.85 & 35.32 & 35.97 & 47.75 \\ 
Pho.     & 42.20 & 40.43 & 39.63 & 41.13 & 25.59 \\ 
Pose     & 34.42 & 32.86 & 31.92 & 33.17 & 23.35 \\ 
Pur.     & 33.59 & 32.50 & 32.22 & 32.71 & 26.43 \\ 
Sit      & 45.03 & 42.33 & 43.56 & 44.26 & 58.61 \\ 
SitD.    & 52.53 & 49.69 & 48.96 & 51.93 & 50.29 \\ 
Sm.      & 37.37 & 36.88 & 36.36 & 37.42 & 39.72 \\ 
Wait     & 33.80 & 32.47 & 32.58 & 32.84 & 24.73 \\ 
WD.      & 37.75 & 36.13 & 34.42 & 35.58 & 27.81 \\ 
Walk     & 25.64 & 25.02 & 23.85 & 25.23 & 25.22 \\ 
WT.      & 27.80 & 26.34 & 25.10 & 26.56 & 23.93 \\ 
\midrule
\textbf{Avg.} & 36.53 & 35.12 & 34.48 & 35.64 & 32.64 \\ 
\bottomrule
\end{tabular}
\end{table*}}

\begin{figure}[!ht]
    \centering
    \includegraphics[width=0.8\linewidth]{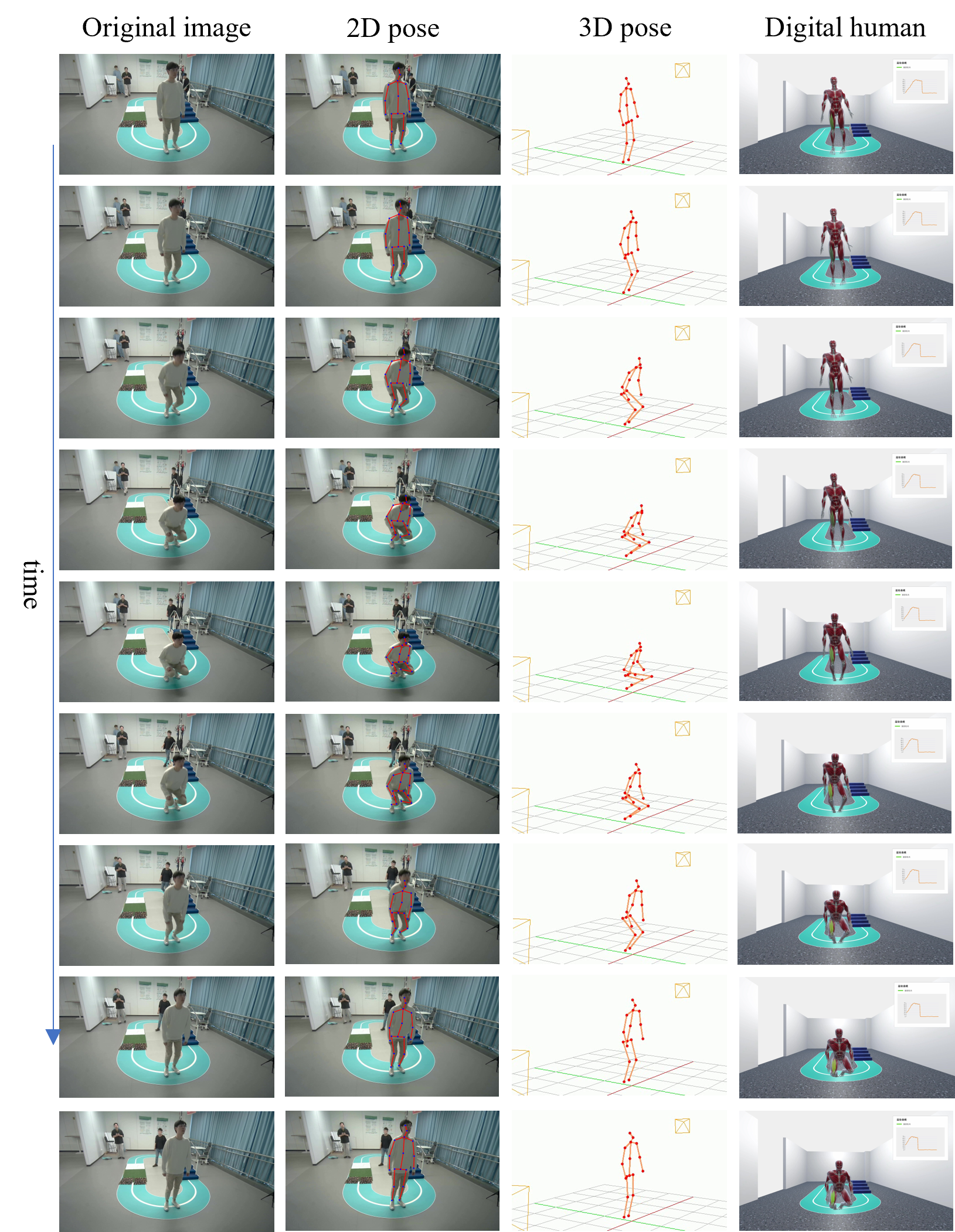}
    \caption{The visualization of input images (only 1 camera), 2D pose, 3D pose generated by our pipeline, and digital human with Color representation of stressed muscles.}
    \label{vis1}
\end{figure}

\begin{figure}[!ht]
    \centering
    \includegraphics[width=0.9\linewidth]{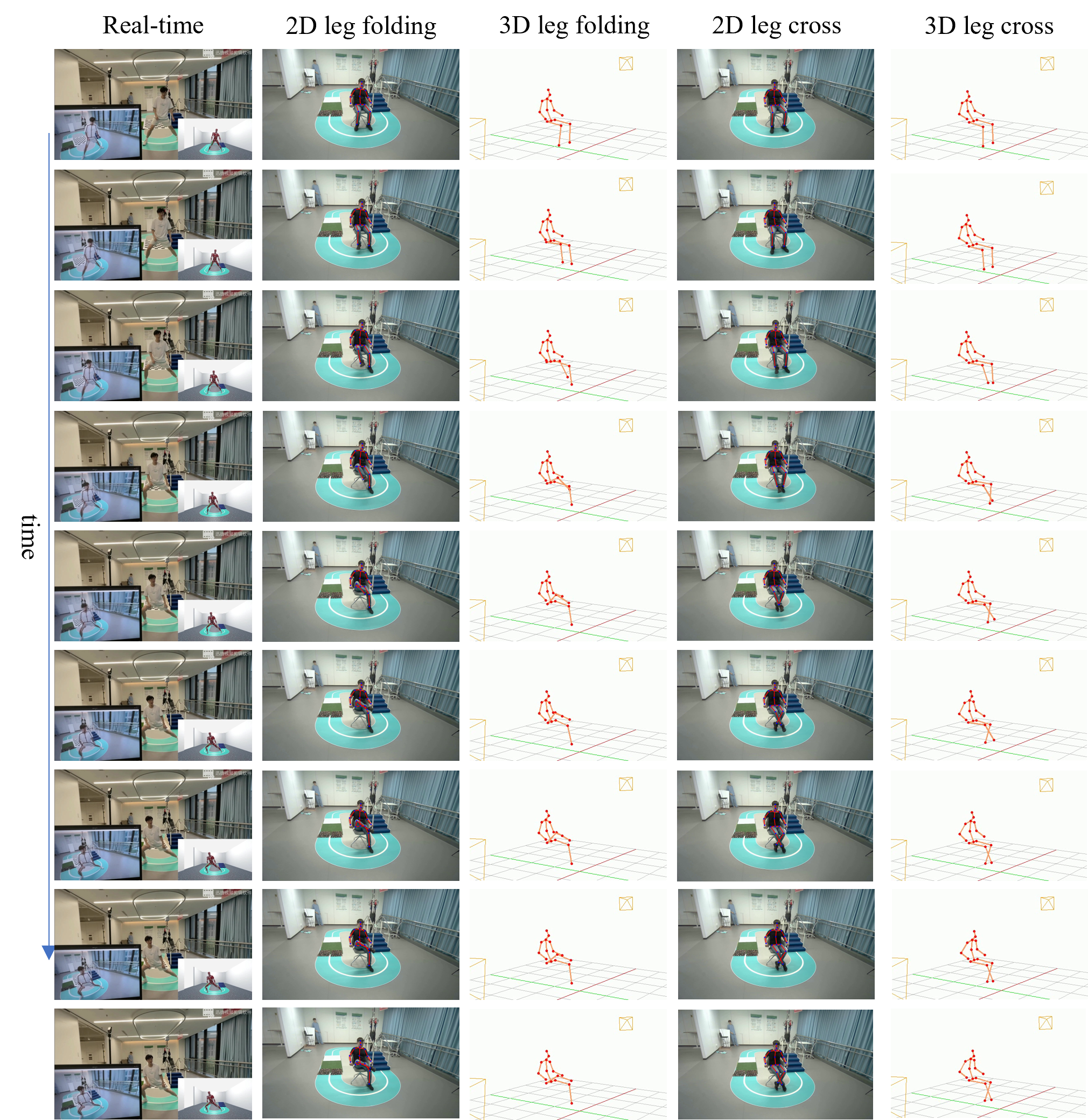}
    \caption{Real-time performance visualization. and 2D Pose and 3D Pose visualization in Leg Fold and Crossing cases.}
    \label{vis2}
\end{figure}

\begin{figure}[!ht]
    \centering
    \includegraphics[width=0.65\linewidth]{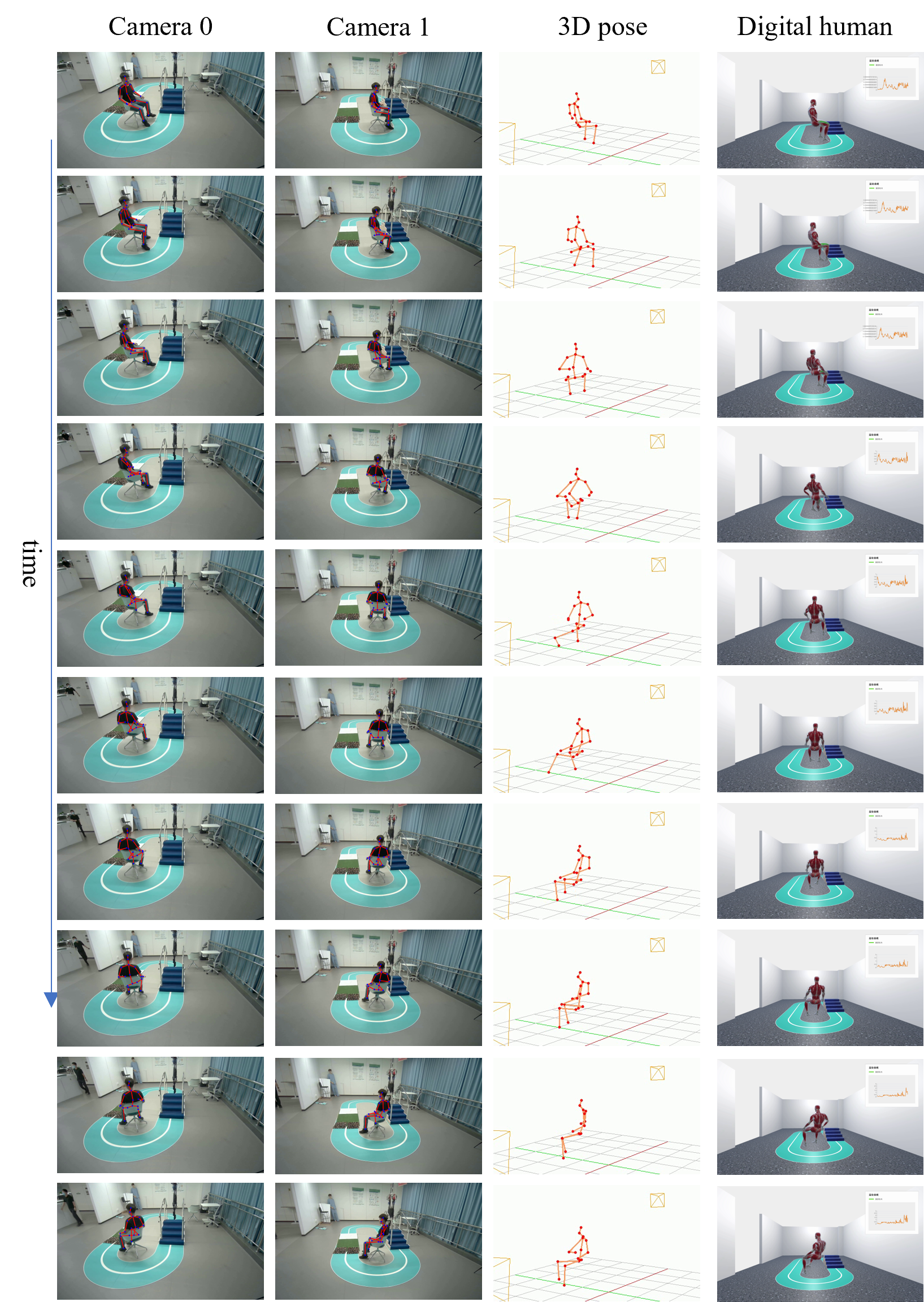}
    \caption{Visualization of occlusion situations by different cameras, and the 3D articulation points and digital humans they generate.}
    \label{vis3}
\end{figure}

Table \ref{tab1} documents the number of different cameras utilized by our model as well as the inference speeds in video testing and real-time scenarios. It is evident that our pipeline delivers commendable real-time performance. ``Num cam'' refers to the number of cameras employed, ``video'' denotes whether the input data is ``offline or online'', indicative of real-time processing. ``Prep in detection'' refers to the image transformation prior to target detection, ``prep in 3d pose'' pertains to image cropping preceding the 3D segment, and ``W/O or W in Fps'' signifies whether fps is doubled through SmoothNet. The rationale for offline being faster than online is attributed to the considerable time and computational resources required for camera reading.

\subsection{Main Results}

We have tested and report the accuracy of our pipeline on the human3.6 dataset. Although our principal innovation is not in pose estimation, our pipeline still demonstrates impressive performance. As illustrated in Tables \ref{tab2} and \ref{tab3}, where * signifies the utilization of ground truth detection boxes provided by human3.6. ``Pipeline'' denotes inference using an end-to-end approach, and $\dagger$ signifies the employment of our in-house developed object detection and tracking algorithm, accelerated through TensorRT, as well as the lightweight 16-bit floating-point model described in Section 4.2. Even without the use of ground truth detection boxes from human3.6, our pipeline exhibits performance comparable to current advanced models under Protocol 1 standards. Under the evaluation criteria of Protocol 2, our pipeline even achieves the best outcomes. This indicates that our pipeline maintains model accuracy while ensuring inference speed, thus validating the effectiveness and practicality of our approach.

\subsection{Visualization}
To demonstrate the efficacy and versatility of our pipeline, we have presented its application in real-world scenarios as shown in Figure \ref{vis1}. At the initial stage of the original image, we display only the perspective of the first camera. Following the object detection, object tracking, and 2D pose estimation modules, we obtained the 2D keypoints. Subsequently, by processing these 2D keypoints from multiple cameras through the 3D pose estimation and the Improved SmoothNet modules, we acquired the 3D keypoints. Ultimately, through the Calculation of Muscle Stress Quantification and Muscle Force Speed computations, we derived the displayed digital human. In practice, the muscle changes in the digital human can be dynamically displayed in real-time through color and motion trajectories. This feature can assist patients in understanding whether they are engaging the correct muscles during exertion or compensating with other muscles. Thus, our pipeline adds value to rehabilitation assistance.

In Figure \ref{vis2}a, we showcased the real-time performance of our proposed pipeline. For actions that are prone to misinterpretation, such as crossed legs (shown in Figure \ref{vis2}b), and for situations with obstructions (shown in Figure \ref{vis2}c), our pipeline demonstrates robust performance. These real-world test results affirm that the proposed pipeline is not only highly accurate but also practical and versatile.
\section{Future works}
\label{sec:futureworl}
While our pipeline has achieved commendable results in Real-life scenarios, there are still areas that require improvement. Our future research will primarily focus on:

\vspace{0.3pt}\noindent\textbf{Occlusion Issues.} When a doctor or family member is assisting in training, occlusions are inevitable, which can be critical for image-based pose estimation, as shown in Figure \ref{vis3}. We plan to address this issue by utilizing a multi-camera collaboration, where the pipeline will automatically select the camera view that is not obstructed for pose estimation.

\vspace{0.3pt}\noindent\textbf{Video Processing.} Currently, our pipeline processes each frame individually and then applies SmoothNet filtering. This does not fully leverage the temporal information in videos. Therefore, we will explore how to extract temporal features without compromising inference speed.

\vspace{0.3pt}\noindent\textbf{Patient-Specific Posture Prior.} Since there is a significant difference between the postures of patients and healthy individuals, the prior information about patients' behavior is very important. Hence, we will focus on data collection and training specific to patients to reinforce our pipeline.

Overall, our pipeline has performed exceedingly well for real-time applications, but we believe that further research in these directions will enhance its accuracy, efficiency, and applicability in medical rehabilitation scenarios.

\section{Conclusion}
\label{sec:conclusion}

We have proposed an end-to-end real-time 3D human pose recognition and motion analysis framework for rehabilitation training. This comprehensive pipeline integrates target detection, high-speed tracking, 2D/3D pose estimation, and temporal refinement via a modified SmoothNet, culminating in muscle stress quantification and digital human visualization. The proposed method not only achieves state-of-the-art accuracy on the Human3.6M dataset but also demonstrates exceptional computational efficiency suitable for real-time deployment. It provides patients and clinicians with a powerful tool for monitoring muscle engagement and ensuring exercise safety through immediate feedback. Beyond its technical contributions, this research holds significant social impact; by facilitating accessible home-based rehabilitation, it has the potential to reduce healthcare costs and alleviate the burden on clinical resources in an aging society. Future work will focus on further optimizing the system's portability for mobile devices and conducting extensive clinical validations to support a wider range of recovery needs.

\clearpage

\bibliographystyle{plainnat}
\bibliography{main}

\end{document}